\documentclass[conference]{IEEEtran}
\IEEEoverridecommandlockouts
\usepackage{cite}
\usepackage{amsmath,amssymb,amsfonts,amsthm}
\usepackage{algorithmic}
\usepackage{graphicx}
\usepackage{textcomp}
\usepackage{xcolor}
\usepackage{mathrsfs}
\usepackage{amsmath}
\usepackage{algorithm}
\usepackage{caption}
\usepackage{subfigure}
\usepackage{float}
\usepackage{hyperref}
\makeatletter

\newcommand{\Rmnum}[1]{\expandafter\@slowromancap\romannumeral #1@}
\makeatother

\def\BibTeX{{\rm B\kern-.05em{\sc i\kern-.025em b}\kern-.08em
    T\kern-.1667em\lower.7ex\hbox{E}\kern-.125emX}}
\begin{document}

\title{Dependency-Aware CAV Task Scheduling via Diffusion-Based Reinforcement Learning\\
}

\author{\IEEEauthorblockN{Xiang Cheng\IEEEauthorrefmark{1}\IEEEauthorrefmark{2}, Zhi Mao\IEEEauthorrefmark{1}, Ying Wang\IEEEauthorrefmark{2}, and Wen Wu\IEEEauthorrefmark{1}}
\IEEEauthorblockA{{\IEEEauthorrefmark{1}Frontier Research Center, Pengcheng Laboratory, Shenzhen, China} \\
\IEEEauthorrefmark{2}School of Information and Communication Engineering, Beijing University of Posts and Telecommunications \\
Email: \{chengx01, maozh, wuw02\}@pcl.ac.cn\IEEEauthorrefmark{1}, and wangying@bupt.edu.cn\IEEEauthorrefmark{2}}}

\maketitle

\begin{abstract}
In this paper, we propose a novel dependency-aware task scheduling strategy for dynamic unmanned aerial vehicle-assisted connected autonomous vehicles (CAVs). Specifically, different computation tasks of CAVs consisting of multiple dependency subtasks are judiciously assigned to nearby CAVs or the base station for promptly completing tasks. Therefore, we formulate a joint scheduling priority and subtask assignment optimization problem with the objective of minimizing the average task completion time. The problem aims at improving the long-term system performance, which is reformulated as a Markov decision process. To solve the problem, we further propose a diffusion-based reinforcement learning algorithm, named \underline{S}ynthetic \underline{D}DQN based \underline{S}ubtasks \underline{S}cheduling, which can make adaptive task scheduling decision in real time.  A diffusion model-based synthetic experience replay is integrated into the reinforcement learning framework, which can generate sufficient synthetic data in experience replay buffer, thereby significantly accelerating convergence and improving sample efficiency. Simulation results demonstrate the effectiveness of the proposed algorithm on reducing task completion time, comparing to benchmark schemes.

\end{abstract}

\section{Introduction}
With advancements in communication and autonomous driving technologies, connected autonomous vehicles (CAVs) have become increasingly prevalent, catering to people's traffic demands \cite{10097602}. They have to execute various computation-intensive and delay-sensitive tasks, including perception fusion, real-time navigation based on video or Augmented Reality (AR), and multimedia entertainment, etc \cite{10158716}. These tasks necessitate joint processing to guarantee safe driving while satisfying the quality of service \cite{10413956}.

Due the limited computing resources for CAVs, it will inevitably prolong tasks completion time when multiple tasks need to be processed simultaneously. To minimize the task completion time, some researches \cite{10521565}\cite{10225421} used mobile edge computing (MEC) to directly offload entire task to the base station (BS) for fast processing through the vehicle-to-infrastructure (V2I) link, which may cause extended completion time due to the additional task transmission. Thus, some literature \cite{10234624}\cite{10225386} had proposed the partial offloading scheme, one part of the task is assigned to the local vehicle, while the remaining is offloaded to edge servers. Although the scheme reduces transmission delay, it remains challenging to guarantee efficient tasks scheduling due to the high dynamic. Consequently, vehicle edge computing (VEC) based tasks scheduling is gradually emerging, other vehicles with available computing resources are acted as extensions of the edge and termed service vehicles (SVs) \cite{10234627}, the tasks generated by task vehicles (TVs) can be offloaded to nearby service vehicles.

Furthermore, the fine-grained vehicular tasks partition and scheduling accelerates the completion of the task \cite{9850402}\cite{10234628}, since tasks can be divided into several dependent subtasks, which are modelled as the directed acyclic graph (DAG) \cite{7463066} describing subtasks interdependency, and offloaded to other SVs or BS. The existing researches illuminate the advantages of the internet of vehicle (IoV) combined with VEC task offloading. However, due to geographical limitations, placing many BSs on highway may not be feasible on economic benefits \cite{9082162}. Thus, for tasks with diverse computation requirements, optimizing the tasks scheduling with high-mobility SVs and the limited BS servers is a challenging problem.

In this paper, we investigate the task scheduling problem for highway CAVs. TVs offload subtasks to SVs with available computing resources, but time-varying task computation requirements and computing resources make it challenging to make adaptive task scheduling decisions in real-time, which will prolong the completion time when the required computing resources of a subtask exceeds the computing capacity of SVs. Due to the advantages of flexibility and easy deployment, the unmanned aerial vehicle (UAV) can be deployed for relaying subtasks to surrounding BS server, aiming at compensating for the offloading needs of TVs when the number of SVs is insufficient or subtasks have a higher workload. Firstly, with the goal of on-demand tasks scheduling, we construct a task scheduling model, i.e., two-side priority adjustment, for determining the scheduling priority of subtasks while considering mobility and resources for selecting optimal offloading targets. Second, we formulate the long-term subtasks scheduling problem that minimizes the average completion time of overall tasks, and the deep reinforcement learning (DRL)-driven synthetic DDQN based subtasks scheduling algorithm (SDSS) is proposed for solving the problem in a dynamic environment. Thirdly, simulation results demonstrate the effectiveness of the proposed algorithm on reducing task completion time. The main contributions of this paper are summarized as follows:
\begin{itemize}
\item We design a dependency-aware task scheduling strategy to reduce task completion time for CAV networks;
\item We formulate a long-term optimization problem for minimizing the overall tasks average completion time and then reformulate it into a Markov decision process (MDP);
\item We propose the SDSS algorithm, which integrates RL and diffusion model to make adaptive task scheduling decisions in real time.
\end{itemize}

The remainder of this paper is organized as follows. We first present the system model and optimization problem in Section II. Section III presents the SDSS algorithm design in detail. The results of the simulation experiments are then presented and analyzed in Section IV. We finally conclude the whole paper in Section V.

\section{System Model and Problem Formulation}
We consider a UAV-assisted highway CAV task scheduling scenario, as depicted in Fig. 1, where a set of TVs can offload the generated computation-intensive tasks that composed of several dependent subtasks to SVs directly via vehickle-to-vehicle (V2V) links, or offload the task to the BS server with the help of the relay UAV, for satisfying TVs tasks offloading demands as needed. In addition, the set of TVs and SVs is denoted as $\mathcal{N}$ and $\mathcal{S}$, respectively.

\subsection{Communication Model}
By allocating the communication resources to different transmitting vehicles, the orthogonal frequency-division multiple access technique \cite{8985313} is considered to avoid interference among multiple vehicles. There are two types of links considered in this paper: V2V links and UAV-assisted V2I links. The TVs can offload tasks using no more than one link type. In V2V links, the data transmission rate for tasks offloading from TV $n$ to SV $s$ can be calculated as
\begin{equation}
\setlength{\abovedisplayskip}{5.5pt}
\setlength{\belowdisplayskip}{1pt}
r_{n,s} = B_{n,s}\log_{2}\left(1+\frac{P_nh_{n,s}^2}{\sigma_n^{2}}\right) , \end{equation}
\\ where ${B_{n,s}}$, $P_{n}^t$, $h_{n,s}$, and $\sigma_n^{2}$ represent the channel bandwidth, transmitting power at TV $n$, channel gain for the link between TV $n$ and SV $s$, and the noise power, respectively.

In V2I links, the UAV can relay the tasks from TV $n$ to the BS server \emph{j}. The data rate from TV $n$ to the UAV, and from UAV to the BS \emph{j} can be calculated as follows,
\begin{equation}r_{n,u} = B_{n,u}\log_{2}\left(1+\frac{P_{n}h_{n,u}^2}{\sigma_n^{2}}\right),
\end{equation}

\begin{equation}r_{u,j} = B_{u,j}\log_{2}\left(1+\frac{P_{u}h_{u,j}^2}{\sigma_n^{2}}\right),
\end{equation}
where $B_{n,u}$ and $B_{u,j}$ represent the channel bandwidth of the links from TV $n$ to UAV $u$, and from UAV $u$ to BS server $j$, respectively. $P_{u}$ is the transmitting power at UAV $n$, $h_{n,u}$ and $h_{uj}$ are the channel gain from $n$ to $u$ and from $u$ to $j$.
\subsection{Task Dependency Model}
\vspace{-0.2em} 
The TV generates \emph{M} consecutive and computation-intensive tasks simultaneously.
\begin{figure}[t]
  \centerline{\includegraphics[width=1\linewidth]{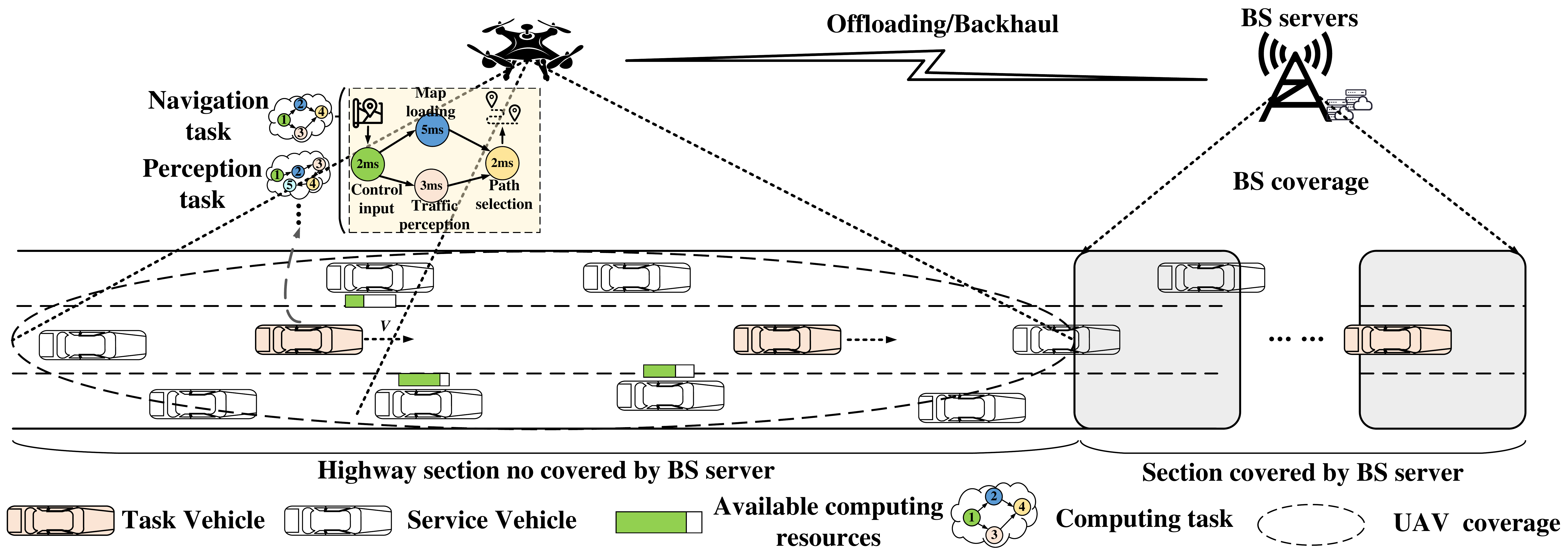}}
  \caption{\ System scenario for UAV-assisted highway CAVs task scheduling.}
  \label{fig}
\end{figure}
Let $\Phi_m$ = ($\omega_m$, $\tau_m$, $f_m$) represents the characteristic of \emph{m}-th task, $1\leq \emph{m}\leq \emph{M}$, where $\omega_m$, $\tau_m$, and $f_m$ represent the computation workload, the maximal tolerant delay, and the computing resource demand of the \emph{m}-th task, respectively. Fig. 2 shows an example of vehicular task DAG with the perception fusion and AR navigation task. They are composed of multiple interdependency subtasks, the subtask set of task \emph{m} is denoted as $\mathcal{W} = \{1, \cdots, \emph{w},\cdots, \emph{W}\}$.

Multiple subtasks are modelled as the DAG to describe their interdependency, the parameter is $\varphi_{m}^w$ = ($\omega_{m}^w$, $\lambda_{m}^w$), where $\omega_{m}^w$ and $\lambda_{m}^w$ represent the workload size and computing resources required of the $w$-th subtask of task \emph{m}. Fig. 2(b) shows an example of an AR navigation task, the entire task can be divided into six subtasks. Control input as the initial subtask, i.e., Subtask 1, its outputted information is used as the input of Subtask 2 map loading and Subtask 3 traffic perception, and then Subtask 4 path selection outputs the optimal driving path based on the results of Subtask 2 and Subtask 3, while Subtask 5 performs video processing based on the perception results of Subtask 3. Finally, Subtask 6 executes the AR navigation using the results of Subtask 5 and Subtask 4.

\subsection{Task Scheduling Model}
\vspace{-0.1em}
\subsubsection{Scheduling Time}
A series of subtasks can be scheduled from TV \emph{n} to nearby SVs, relayed to BS by UAV, or executed on the local TV. The completion time of each subtask is composed of the transmission delay and computing delay. The transmission delay is ignored when the subtask is executed on a local TV. When subtasks are executed in TV \emph{n} locally, the computing time is calculated as
\begin{equation}T_w^{n}=\lambda_m^w/f_n ,\end{equation}
where $\lambda_m^w$ and $f_n$ are computing resources required to complete $w$ and available computing capacity of the TV \emph{n}. 

When one subtask $w$ is offloaded to SVs, the completion time of the subtask is given by 
\begin{equation}T_w^{s,\text{total}}=T_{w}^{n,s}+T_w^{s}\end{equation} where $T_{w}^{n,s}$ and $T_w^{s}$ represent 
the data transmission and computing delay, i.e., $T_{w}^{n,s}=\omega_{m}^w/r_{n,s}$ and $T_w^{s}=\lambda_{m}^w/f_s$, and the $f_s$ indicates available computing capacity of the SV \emph{s}.
\begin{figure}[t]
\centerline{\includegraphics[width=1\linewidth]{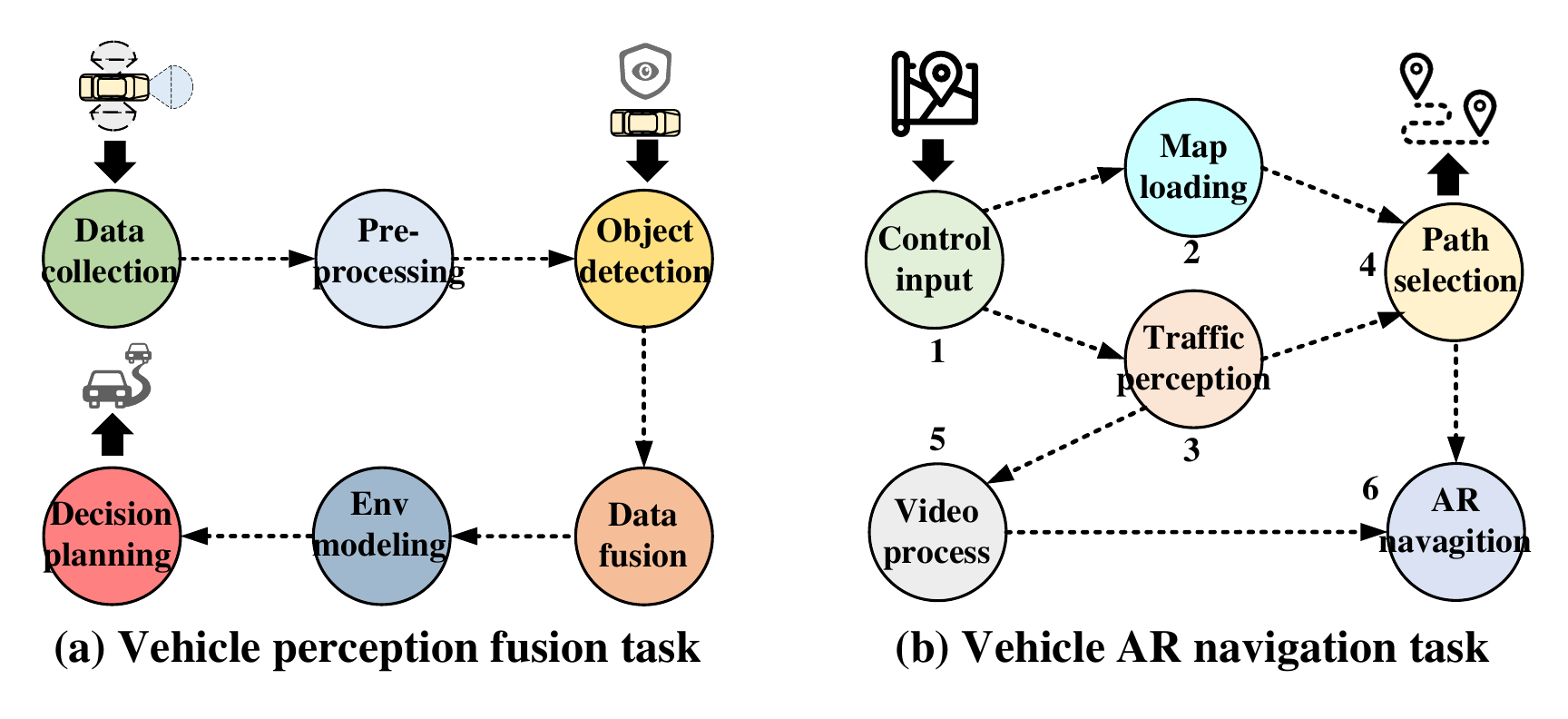}}
\caption{Example of vehicular task DAG.}
\label{fig}
\end{figure}
When the computing demands exceed the computing capacity of the SV, the subtasks could be relayed to BS. The completion time of subtask executed in the BS server $j$ is given as
\begin{equation}T_w^{j,\text{total}}=T_{w}^{n,u,j}+T_w^{j},\end{equation} where $T_{w}^{n,u,j}$ and $T_w^{j}$ represent the transmission delay from the SV to BS and computing delay. Then we have $T_{w}^{n,u,j}=\omega_{m}^w/r_{n,u}+\omega_{m}^w/r_{u,j}$ and $T_w^{j}=\lambda_m^w/f_j$, where $f_j$ indicates the available computing capacity of the BS server \emph{j}. 

\subsubsection{Scheduling Strategy}
For dependent subtasks, successor subtasks can only be executed after completing the predecessor ones. Due to the dynamic network topology and time-varying computing resources of SVs, it is worth noting when and to whom to offload, which will determine whether it can achieve the lowest completion time. Thus, the scheduling priority of subtasks needs to be arranged while considering the selection of offloading targets. We design a two-side priority adjustment mechanism, aiming at selecting the optimal offloading targets for all subtasks. It includes three parts: 

Subtasks scheduling priority: For serial dependency, the priority of the predecessor subtasks is higher than the successor ones. The scheduling priority of subtask $w$ is denoted as $\alpha_w$, it needs to meet the following constraint
\begin{equation}\alpha_{w}>\alpha_{w+1}, \forall w\in \mathcal{W},\end{equation}
where the successor subtasks only can begin to execute after completing the predecessor ones, i.e., $T_{w+1}^{\mathrm{b}}\geq T_{w}^{\mathrm{e}}$, $T_{w+1}^{\mathrm{b}}$ and $T_{w}^{\mathrm{e}}$ represent the beginning time of successor subtask $w+1$ and the ending time of predecessor subtask $w$, respectively.

Offloading targets selection priority: First, the priorities of SVs are determined by comparing the distance between SV and TV. In addition, the priorities are resorted based on available computing resources $f_s$. Specifically, the SVs with smaller distances and higher computing resources are marked as the highest selection priority.

Two-side priority adjustment: To guarantee the minimum completion time of overall tasks, the priority of offloading targets needs to be adaptively adjusted based on computing demands, dynamic topology, and available resources after determining the scheduling priorities of subtasks. The goal is to minimize the task completion time according to the rapid adjustment of scheduling and offloading orders.

\subsection{Problem Formulation}
In this part, we formulate the optimization objective that minimizes the average completion time of overall tasks. Firstly, the completion time $\tau_m$ of a task \emph{m} composed of multiple dependency subtasks is denoted as follows,
\begin{equation}
\setlength{\abovedisplayskip}{1pt}
\setlength{\belowdisplayskip}{4pt}
\tau_m=\sum_{w=1}^W(x_{w,s}T_w^{s,\text{total}}+y_{w,j}T_w^{j,\text{total}}+z_{w,n}T_w^n) ,\end{equation}
where the V2V scheduling of subtask $w$, the V2I scheduling, and the local execution are denoted by the binary variable $x_{w,s}$, $y_{w,j}$, and $z_{w,n}$, which are equal to 1 if the V2V scheduling, UAV-assisted V2I scheduling, and local execution are used, respectively, otherwise, 0. It is assumed that a subtask $w$ can be executed at only one target: local TV \emph{n}, one SV \emph{s}, or BS server \emph{j}. The Eq. (9) must be met to guarantee that each subtask can only be offloaded to one target,
\begin{equation}{x_{w,s_i} +y_{w,j}+z_{w,n}=1.}
\end{equation}

Therefore, considering the scheduling and offloading of subtasks, the problem of minimizing the overall average completion time of all subtasks can be formulated as
\begin{subequations}\begin{align}\mathcal{P}:\min\limits_{\alpha_{w}, \ x_{w,s}, \ y_{w,j}, \ z_{w,n}}&~ \frac{1}{M}\sum_{m=1}^{M}\sum_{w=1}^{W}\tau_{m,w} \\
\text{s.t.} &~\lambda_{m}^{w}\leq f_{s},\forall w\in\mathcal{W},\forall s\in \mathcal{S}, \\ &~\sum_{w=1}^{W}\tau_{w}\leq\tau_{m},\forall m, \\  &~T_{w+1}^{\mathrm{b}}\geq T_{w}^{\mathrm{e}},\\  &~x_{w,s}+y_{w,j}+z_{w,n}=1, \\  &~x_{w,s},\ y_{w,j},\ z_{w,n}\in \{0,1\},  \\  &~\alpha_{_{w}}\in\mathbb{N},\end{align}\end{subequations} where $\alpha_w$, $x_{w,s_i}, y_{w,j}, z_{w,n}$ represent optimization variables related to two-side priority adjustment, $\alpha_w$ is the scheduling decision of subtasks, and $x_{w,s_i}, y_{w,j}, z_{w,n}$ represent the offloading decision. Constraint (10b) states the computing resources of the selected SV cannot be smaller than the required computing resource of subtask $w$. (10c) guarantees the completion time of the task does not exceeds the maximal tolerant delay. (10d) means that the successor subtasks can only be executed after completing the predecessor subtasks. (10e) ensures that a subtask can only be assigned to one target. (10f) and (10g) defines the range of variables.

Since the above optimization problem $\mathcal{P}$ is nonlinear and has multiple integer variables, it is difficult to find the optimal solution within polynomial time under the dynamic subtasks workload, network environment, and computing resources of SVs. Moreover, the selection of offloading targets also needs to consider the status of each SV and subtasks workload. The above factors make it difficult to find the optimal scheduling strategy using traditional optimization methods for the optimization problem. Therefore, we propose the data-driven DRL algorithm \cite{sutton2018reinforcement} to solve the above problem. The key idea is to model the long-term scheduling process of dependent subtasks in the dynamic environment as an MDP.

\section{Algorithm Design}
In this section, we design the SDSS algorithm based on synthetic experience replay (SER) \cite{lu2024synthetic}-double deep Q network (DDQN) \cite{van2016deep}, where DDQN is the representative DRL algorithm in discrete decisions, the novel combination aims to adaptively control the discrete scheduling decisions while guaranteeing the efficient strategies exploration by additionally generating high-reward transitions from the diffusion-based SER module. We describe the long-term subtasks scheduling and offloading problem as an MDP and then propose the SDSS algorithm to solve it.

\subsection{MDP Formulation}\label{}
The optimization problem $\mathcal{P}$ is described as an MDP four-tuple $\{S,A,P,R\}$, including state \emph{S}, action \emph{A}, transition probability \emph{P}, and reward \emph{R}. The main components are as follows.

\subsubsection{State} The state is denoted as the combination of subtasks information, scheduling decisions, and offloading targets information. The state for subtask $w$ is denoted as
\begin{equation}\mathcal{S}=\{I_w,a_{w},O_r\}, \end{equation} where $\emph{I}_w$ denotes information of subtasks, including workload $\omega_{m}^w$ and interdependency, i.e., the indicator of predecessor and successor subtask $\emph{Pre}_w$ and $\emph{Suc}_w$. $a_w$ denotes the scheduling decision of subtasks. $O_r$ is the offloading targets information, including the available computing resources of SVs and BS server $f_s$ and $f_j$, and the distance between TVs and SV.

\subsubsection{Action} Each subtask $w$ of the vehicle task \emph{m} can be computed in local TV \emph{n} or offloaded to SV \emph{s} and BS server \emph{j}. For a whole vehicle task, the scheduling action of all subtasks can be represented as \begin{equation}a_{w}=\{\alpha_w, \mathcal{B}_w\} ,\end{equation} where $\alpha_w$ is the scheduling priority of subtasks and $\mathcal{B}_w=\{x_{w,s},y_{w,j},z_{w,n}\}$ as an array denotes the different selection of offloading targets.

\subsubsection{Reward Function} The reward function is designed as the negative increment of delay after making a scheduling action, it is denoted as
\begin{equation}R\left(s_w, a_w\right)=-\Delta \tau_w ,\end{equation}
where $\Delta \tau_w$ is represented as the difference of completion delay between two adjacent subtasks scheduling decisions, i.e., $\Delta \tau_w=\tau_{w+1}^{a_{w+1}}-\tau_w^{a_{w}}$. The goal is to find the optimal decision to obtain the maximum cumulative reward.
\subsection{SDSS Algorithm}
\begin{figure}[t]
\centerline{\includegraphics[width=1\linewidth]{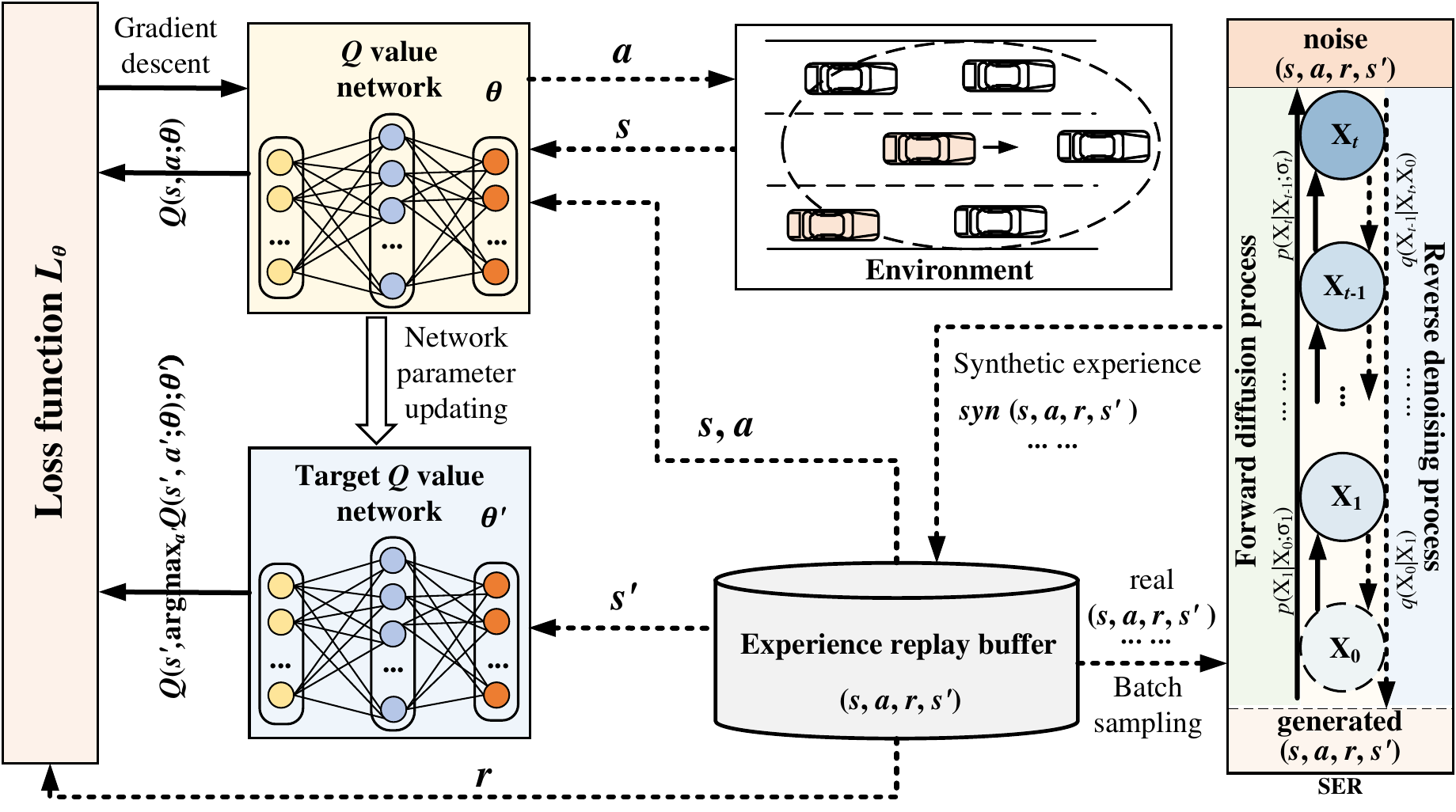}}
\caption{\ Model structure of the SDSS algorithm.}
\label{fig}
\end{figure}
The model structure of the SDSS algorithm based on the Synthtic-DDQN is shown in Fig. 3, the algorithm structure mainly includes \emph{Q} value  network and target \emph{Q} network and an improved experience replay buffer. 

On the one hand, the two networks represent the \emph{Q} value network $\theta$, i.e., policy network and target \emph{Q} value network $\theta^{\prime}$. The policy network estimates the \emph{Q} value of all possible actions $a_{w}$ in state $\emph{S}_w$ and outputs the action \emph{a} corresponding to the maximum \emph{Q} value, the calculation of action selection process is shown as 

\begin{equation}a_{\text{max}}\left(s,\theta \right)=\text{argmax}_{a'}Q_\text{estim}\left(s,a,\theta \right) ,\end{equation}
and then the target \emph{Q} value network $\theta^{\prime}$ uses the action with the maximum \emph{Q} value to calculate the expected value of the next state $\emph{S}_{w+1}$, the calculation of the target \emph{Q} value is denoted as
\begin{equation}y_{\text{target}}=r+\gamma Q\left(s^{\prime},\text{argmax}_{a^{\prime}}Q_\text{estim}\left(s^{\prime},a,\theta \right),\theta^{\prime}\right) ,\end{equation}
where $\gamma$ represents the discount factor, which can be used to adjust the long-term reward, and its range is set is $\gamma \in \left(0.1,\ 0.99\right)$. Subsequently, it can obtain the loss function by comparing the target \emph{Q} value, i.e., $y_{\text{target}}$ with the output $Q_\text{estim}\left(s,a,\theta \right)$ of \emph{Q} network, 
\begin{equation}
  \begin{aligned}
L\left(\theta\right)= & \mathbf{E}_{s,a;s^{\prime}}\{\frac{1}{2}\left(r+\gamma Q\left(s^{\prime},\text{argmax}_{a^{\prime}}Q_\text{estim}\left(s^{\prime},a,\theta \right);\theta^{\prime}\right) \right. \\
& \left.-Q_\text{estim}\left(s,a;\theta \right)\right)^{2}\},
  \end{aligned}
\end{equation}the above operation of decoupling action selection and value estimation can effectively avoid the overestimation problem of the target \emph{Q} value, which is helpful to improve the reliability of the high dimension discrete strategies.

On the other hand, the main goal placing the experience replay buffer between the networks and the environment interaction is to store the transitions obtained from the agent's interaction with the environment. These stored transitions $\left(s,a,r,s^{\prime}\right)$ , as the exploration experience, can provide training data for these networks, which also makes the training samples of the agent have a certain diversity. 

In practice, there exist a large number of ineffective transitions in the initial strategy exploration stage. It  will inevitably cause the difficulty in convergence by sampling and utilizing these data for agent training. Therefore, to improve the algorithm convergence ability and sample efficiency, we consider using the emerging generative diffusion model \cite{10419041} to provide real and synthetic transitions for agent training through diffusion-based transitions generation, i.e., the SER module. Two steps of SER module are as follows.

\subsubsection{The forward noising} For a original data distribution $p\left(\mathbf{x}\right)$ with standard deviation $\sigma$, $\mathbf{x}$ represents the real transitions data, considering the noised distribution $p(\mathbf{x};\sigma)$ that obtained by adding independent and identically distributed Gaussian noise of deviation $\sigma$ to the $p\left(\mathbf{x}\right)$, i.e., $p\left(\mathbf{x}_t|\mathbf{x}_0; \sigma_n \right)$, which will make the final data distribution with unknown noise become the indistinguishable random noise.

\subsubsection{The reverse denoising} The essential of denoising process is to learn iteratively to reverse forward noising process and generate samples from unknown noise, i.e., $q(\mathbf{x_t}|\mathbf{x_{t-1}};\mathbf{x_0})$. For the training of the denoising model, it begins with the training of DDQN, and applying the score function based ordinary differential equation solver \cite{karras2022elucidating} to perform the denoising process.

The Alg. 1 is designed to solve the optimization problem of the subtasks scheduling for continuous CAVs task.
\vspace{-0.6em}
\begin{algorithm}[t]
    \caption{Proposed SDSS algorithm}
    \label{alg:AOS}
    \renewcommand{\algorithmicrequire}{\textbf{Input:}}
    \renewcommand{\algorithmicensure}{\textbf{Output:}}
   
    \begin{algorithmic}[1]
        \REQUIRE The subtask info $I_w$, offloading target info $O_r$, data ratio $r$, and discount factor $\gamma$;\\
        \ENSURE The estimation values of two \emph{Q} networks;\\
        \STATE Initialize experience replay buffer $\mathcal{D}$, denoising model, networks parameter $\theta$ and $\theta^{\prime}$;
        \FOR{episode = 1: \emph{M}} 
            \STATE episode = 1,  Reset the environment and initialize state space $\mathcal{S}$;
            \FOR{step = 1: \emph{T}} 
            \STATE step = 1, agent outputs actions $a$, then selects action and executes, $\text{argmax}_{a'}Q_\text{estim}\left(s,a,\theta \right)$;
            \STATE Complete subtask scheduling and offloading target selection while calculating $r_w$, and $S_w\leftarrow S_{w+1}$;
            \STATE Store the transitions $\left(s,a,r,s^{\prime}\right)$ into $\mathcal{D}$;
            \STATE The real transitions from $\mathcal{D}$ are sampled into SER module for updating the forward diffusion process;
            \STATE Generate samples from reverse denosing process by diffusion step iteratively, and adding to $\mathcal{D}$, $\mathcal{D}\leftarrow \mathcal{D}_{syn}$;
            \STATE Train the agent makes scheduling policy by sampling from $\mathcal{D}$ with ratio \emph{r};
            \STATE Calculating the target \emph{Q} value using Eq. (15);
            \STATE The \emph{Q} network $\theta$ is updated by gradient descent of loss function Eq. (16); 
            \STATE step + 1
            \ENDFOR
          \STATE episode + 1
        \ENDFOR
        
    \end{algorithmic}
\end{algorithm}
\vspace{-0.6em}
\section{Performance Evaluation}
\addtolength{\topmargin}{0.05in} 
In this section, simulation results are provided to demonstrate the effectiveness of the proposed SDSS algorithm. The traffic simulator Simulation of Urban MObility (SUMO) \cite{lopez2018microscopic} is used to generate vehicle mobility. The DAG generator \cite{8847369} is used to simulate task DAG with different dependencies. We consider the simulation experiment of CAV tasks scheduling in a section of one km highway lacking BS coverage, where the single relaying UAV can hover to cover the whole section. The simulation parameters and the algorithm hyperparameter configurations are listed in Table \uppercase\expandafter{\romannumeral 1}. Next, we will first compare and analyze the convergence performance of the proposed SDSS algorithm with the original DDQN algorithm \cite{van2016deep} in part of validation of analytical results, and compare the algorithms performance with the random subtasks scheduling scheme under diverse workload and computing capacity conditions in performance comparisons.
\begin{table}[t]
  \centering
  \caption{Simulation parameters.}\label{}
  \begin{tabular}{lc}
    \hline
    \hline
    Parameter & Value \\
    \hline
    Number of TVs and SVs \ $(N,S)$ &  \{2, 5\} \\
    Number of subtasks of single task \ $(W)$ & \{4$\sim$6\} \\ 
    Computing power of TVs, SVs \ $(f_n,f_s)$ & \{2, 2$\sim$8\} GHz \\
    Computing power of BS server \  $(f_j)$ & 50 GHz \\
    Transmit power of vehicle and UAV \ $(P_n,P_u)$ & \{20, 30\} dBm \\ 
    Bandwidth of vehicle and UAV \ $(B_n,B_u)$ & \{5, 10\} MHz \\ 
    Maximum tolerant delay of single task \ $(\tau_m)$ & 650 ms\\
    Workload size of a subtask \ $(\omega_m^w)$ & \{500$\sim$5000\} KB \\
    Length of highway section &  1 km \\
    Learning rate and discount factor & \{0.001, 0.95\} \\
    Experience replay size & 100000 \\
    Denoising step  & 10 \\
    \hline
  \end{tabular}
\end{table}
\subsection{Validation of Analytical Results}
\begin{figure}[t]
\centerline{\includegraphics[width=0.9\linewidth]{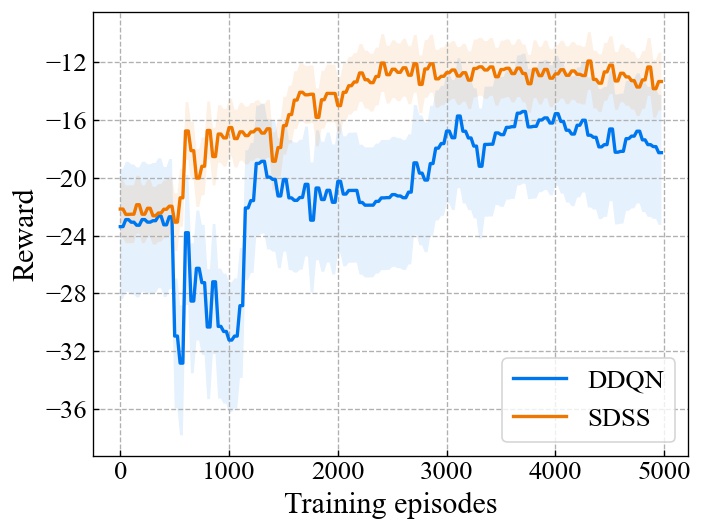}}
\caption{The algorithm convergence curve.}
\label{fig}
\end{figure}

Figure 4 shows the reward convergence curve of the SDSS algorithm compared with DDQN, where the curve is the mean value with five random seed simulations, and the shadow part represents their positive and negative standard deviation. It is obvious that the proposed SDSS algorithm can achieve rapid convergence and higher reward, and the stability of the algorithm is better than the original DDQN. The main reason is that SDSS can train the agent with generated experience data under the original strategy exploration stage. When the agent begins to learn the discrete decision of subtasks offloading, the generated transitions from the real interaction experiences can help it avoid the inefficient trial and error process, which not only can reduce the train time but also improve the performance of discrete decisions.


\subsection{Performance Comparisons}
In this part, the average completion time of the proposed SDSS algorithm is compared with that of the DDQN and random offloading schemes. We consider evaluating algorithm performance under different subtasks workload size and SV computing capacity, i.e., 1) subtasks workload size (Unit: KB): four subtasks and three combinations with low, high, and mix workload size, i.e., W1 (500, 600, 700, 800), W2 (4000, 4300, 4600, 4900), and W3 (800, 2500, 1200, 4500). 2) SVs computing capacity (Unit: GHz): five SVs and three combinations with low, middle, and high available computing resources, i.e., C1 (2, 3, 4, 3, 2), C2 (4, 5, 5, 8, 3), and C3 (6, 7, 5, 8, 8).
\begin{figure}[t]
  \centering
    \begin{minipage}[b]{\linewidth}
      \subfigure[Delay \emph{vs} subtasks workload size]{
        \includegraphics[width=0.46\linewidth]{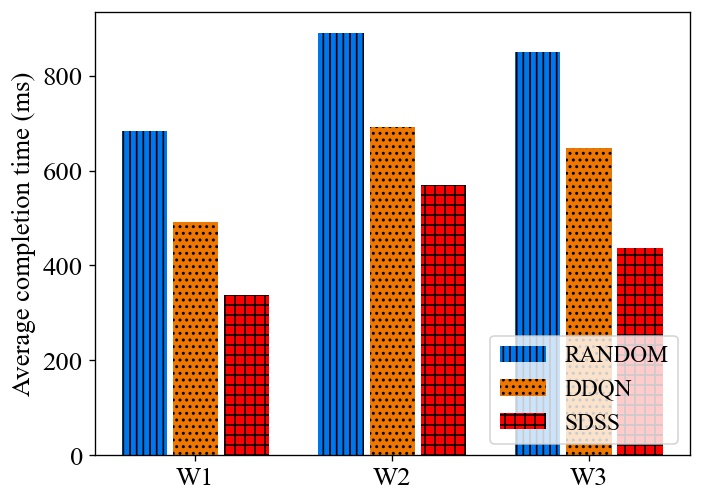}
      }
      \subfigure[Delay \emph{vs} SVs computing capacity]{
        \includegraphics[width=0.46\linewidth]{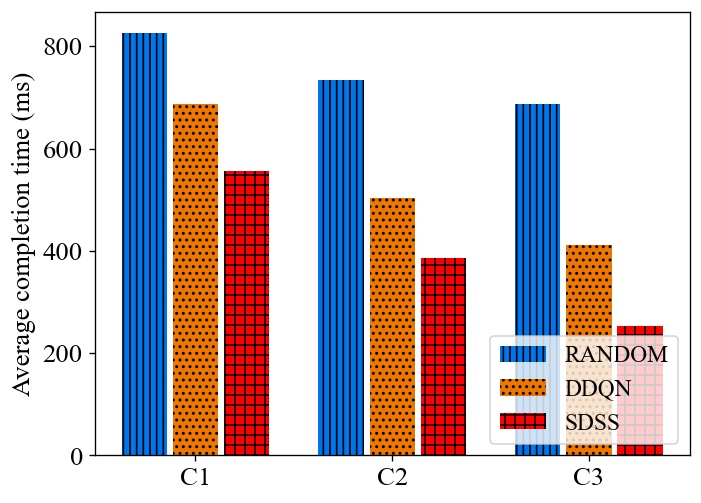}
      }
    \end{minipage}
    \caption{The performance comparison.}
    \label{}
  \end{figure}

Figure 5 indicates the performance comparison with two schemes under different subtasks workload size and SVs computing capacity. In Fig. 5(a), we compare the average completion time of different algorithms. It can be seen that the proposed SDSS algorithm can achieve a lower task completion time under different computing requirements, which demonstrates the higher robustness of scheduling decisions. The main reason is that SDSS can adaptively adjust the offloading target selection, while determining the subtasks scheduling priority. It can achieve the efficient subtasks offloading by estimating the target states, subtask completion time, and overall task completion time.

Figure 5(b) illustrates that subtasks with high workload can be relayed to the BS server through UAV, but the data transmission overhead cannot be ignored. The computing resources of SVs need to be effectively utilized to complete all subtasks within the tolerant delay. Regarding combination C1, due to the low computing resources of SVs, SDSS can choose to offload some subtasks with high workload to BS by UAV, which will lead to a higher completion delay. When existing the SV with high computing resources, SDSS can adaptively make subtasks offloading decisions to reduce overall task average completion delay.
\section{Conclusion}

In this paper, we have studied the task scheduling problem for CAV dependency subtasks offloading in highway scenario. Firstly, the task dependency model and scheduling model have been established to formulate the optimization problem of minimizing the average completion time of overall tasks. Then, the long-term optimization problem is reformulated as an MDP to find the optimal subtask scheduling strategies. Moreover, to achieve the objective, we have designed the SDSS algorithm based on synthetic-DDQN. Finally, we have shown that SDSS can yield faster scheduling decision exploration and lower task completion time in a dynamic environment than other schemes under different scales of subtasks workload size and SVs computing capacity.

\bibliographystyle{IEEEtran}

\bibliography{ref.bib}
\end{document}